\documentclass{article} % For LaTeX2e
\usepackage{nips15submit_e,times}
\usepackage{hyperref}
\usepackage{url}
\usepackage{graphicx}
\usepackage{caption}
\title{Learning Linguistic Biomarkers for Predicting Mild Cognitive Impairment using Compound Skip-grams}

\author{
Sylvester Olubolu Orimaye \\
Intelligent Health Research Group\\
School of Information Technology\\
Monash University Malaysia \\
\texttt{sylvester.orimaye@monash.edu} \\
\And
Kah Yee Tai \\
Intelligent Health Research Group\\
School of Information Technology\\
Monash University Malaysia \\
\texttt{kytai2@student.monash.edu} \\
\AND
Jojo Sze-Meng Wong \\
Intelligent Health Research Group\\
School of Information Technology\\
Monash University Malaysia \\
\texttt{jojo.wong@monash.edu} \\
\And
Chee Piau Wong \\
Jeffrey Cheah School of \\
Medicine and Health Sciences\\
Monash University Malaysia \\
\texttt{wong.chee.piau@monash.edu} \\
}

\nipsfinalcopy % Uncomment for camera-ready version

\begin{document}

\maketitle

\begin{abstract}
Predicting Mild Cognitive Impairment (MCI) is currently a challenge as existing diagnostic criteria rely on neuropsychological examinations. Automated Machine Learning (ML) models that are trained on verbal utterances of MCI patients can aid diagnosis. Using a combination of skip-gram features, our model learned several linguistic biomarkers to distinguish between 19 patients with MCI and 19 healthy control individuals from the DementiaBank language transcript clinical dataset. Results show that a model with compound of skip-grams has better AUC and could help ML prediction on small MCI data sample.

\end{abstract}

\section{Introduction}

MCI is typically diagnosed through neuropsychological examinations with series of cognitive tests \cite{mitolo:2015,goryawala:2015}. For example, the Mini-Mental State Examination (MMSE) and the Montreal Cognitive Assessment (MoCA) screening tools use series of questions to assess different cognitive abilities \cite{roselli:2009}. Since MCI causes deterioration of nerve cells that control cognitive, speech, and language processes \cite{reilly:2010,verma:2012}, linguistic impairments from verbal utterances could indicate signs of MCI \cite{williams:2013,fraser:2014,tillas:2015}. 

In \cite{roark:2011}, complex syntactic features were used to distinguish between 37 patients with MCI and 37 healthy control group. Seven statistically significant `immediate logic memory' linguistic features were combined with several test scores to achieve 86.1\% AUC. In contrast, we distinguished MCI patients using several skip-grams alone \cite{guthrie:2006}. The skip-grams are capable of representing the MCI patients' language space due to the lexical and syntactic errors that are commonly observed in their language. Thus, we introduce an extensive use of word skip-grams to predicting MCI.

\section{Learning linguistic skip-gram biomarkers}

Skip-grams are commonly used in statistical language models for natural language processing (NLP) problems such as speech processing \cite{guthrie:2006}. Unlike the ordinary $n$-grams, word tokens are skipped intermittently while creating the $n$-grams. For instance, in the sentence \emph{``take the Cookie Jar"}, there are three conventional \emph{bigrams}: ``take the", ``the Cookie", and ``Cookie Jar". With skip-gram, one might skip one word intermittently for creating additional bigrams, which include ``take Cookie", and ``the jar". We believe such skip-grams could capture unique linguistic biomarkers in verbal utterances. Thus, we used a compound of skip-grams to effectively distinguish the MCI patients from their healthy control individuals. For each sentence $S = \{w_1 ... w_m\}$ in a verbal dialogue, we define $k$-skip-$n$-grams as a set of $n$-gram tokens $T_{n-gram} = \{w_a,...,w_{a+n-k},...,w_{a+n},...,w_{m-n},...,w_{(m-n)+n-k},...,w_m\}$, where $n$ is the specified $n$-gram (e.g. 2 for $bigram$ and 3 for $trigram$), $m$ is the number of word tokens in $S$, $k$ is the number of word skip between $n$-grams given that $k<m$, and  $a = \{1,...,m-n\}$. Thus for the sentence \emph{``take the Cookie Jar from the cabinet"}, $1$-skip-$2$-grams will give $\{$`take Cookie', `the jar', `Cookie from', `jar the', `from cabinet'$\}$ and $1$-skip-$3$-grams will produce $\{$`take Cookie Jar', `take the Jar', `the Jar from', `the Cookie from', `Cookie Jar the', `Cookie from the', `Jar the cabinet', `Jar from cabinet'$\}$.

\section{Experiments and Results}

The DementiaBank\footnote{\url{https://talkbank.org/DementiaBank/}} clinical dataset was used. The dataset contains English language transcripts of multiple verbal interviews, where MCI and Control participants described the Cookie-Theft picture component of the Boston Diagnostic Aphasia Examination \cite{kaplan:2001}. Thus, we extracted a combination of skip-grams features from the transcript files of the `last' interview with the participants. There are 19 MCI patients with an approximate age range of 49 to 90 years. Similarly, we selected an equivalent 19 healthy control individuals from the dataset with an approximate age range of 46 to 81 years. We identified hyperparameters using Auto-Weka \cite{thornton:2013} for four WEKA\footnote{\url{http://www.cs.waikato.ac.nz/ml/weka/}} ML algorithms (i.e. SVM, Na\"{i}ve Bayes, Decision Trees, and Logistic). A separate validation set of transcript files from the `second to the last' visits to the participants were used to find hyperparameters. The validation set has 8 patients with MCI and Control, respectively, using the top 1000 skip-gram features only. 

As baseline, we implemented all the `7 Wechsler Logical Memory I' significant features in \cite{roark:2011}, which includes Words per clause, Part-Of-Speech cross entropy, content density, Standard Pause Rate, Total Phonation Time, Phonation Rate, and Transformed Phonation Rate. The Wechsler Logical Memory I is a narrative memory task, which required the subjects to listen to a story and then recall everything they can from the story. That task allowed the subjects to formulate original language structures on their own, which helps to captured both linguistic and memory deficiencies from the subjects by using various language and speech measures. Similarly, we believe that the Cookie-Theft picture description task captured various linguistic deficiencies, which can be detected by a compound of skip-grams. Figures 1-5 show the accuracy with different top skip-gram features. Table \ref{tab:auc} compares our models with the baseline using the weighted Precision, Recall, F-measure, and AUC from 10-fold cross-validation on top 200 combined skip-grams.

\begin{figure*}[!htb]
\centering
\minipage{0.30\textwidth}
  \includegraphics[width=40mm,height=40mm]{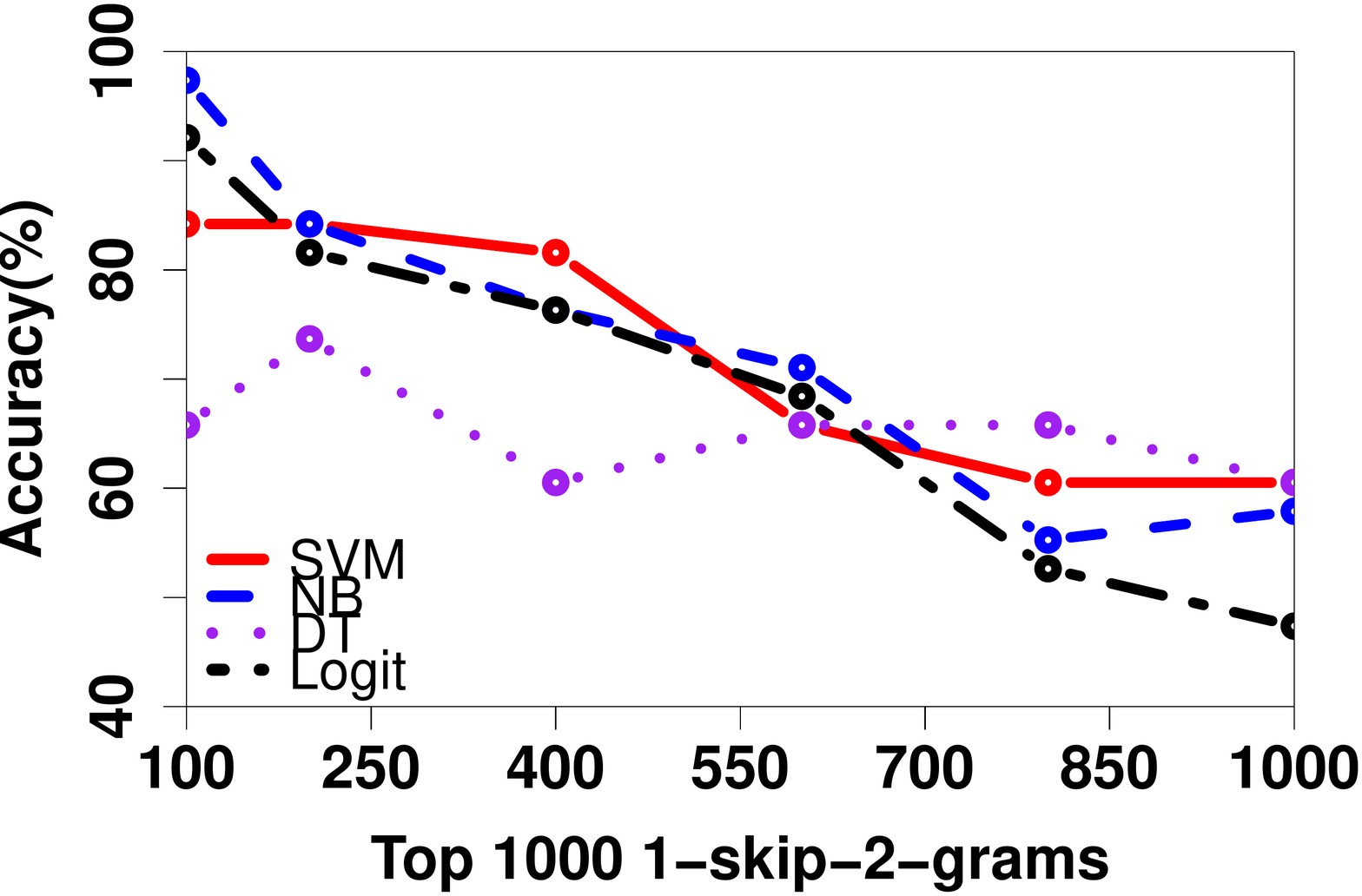}
  \caption{1-skip-2-grams.}\label{fig:1skip2}
\endminipage\hfill
\minipage{0.30\textwidth}
  \includegraphics[width=40mm,height=40mm]{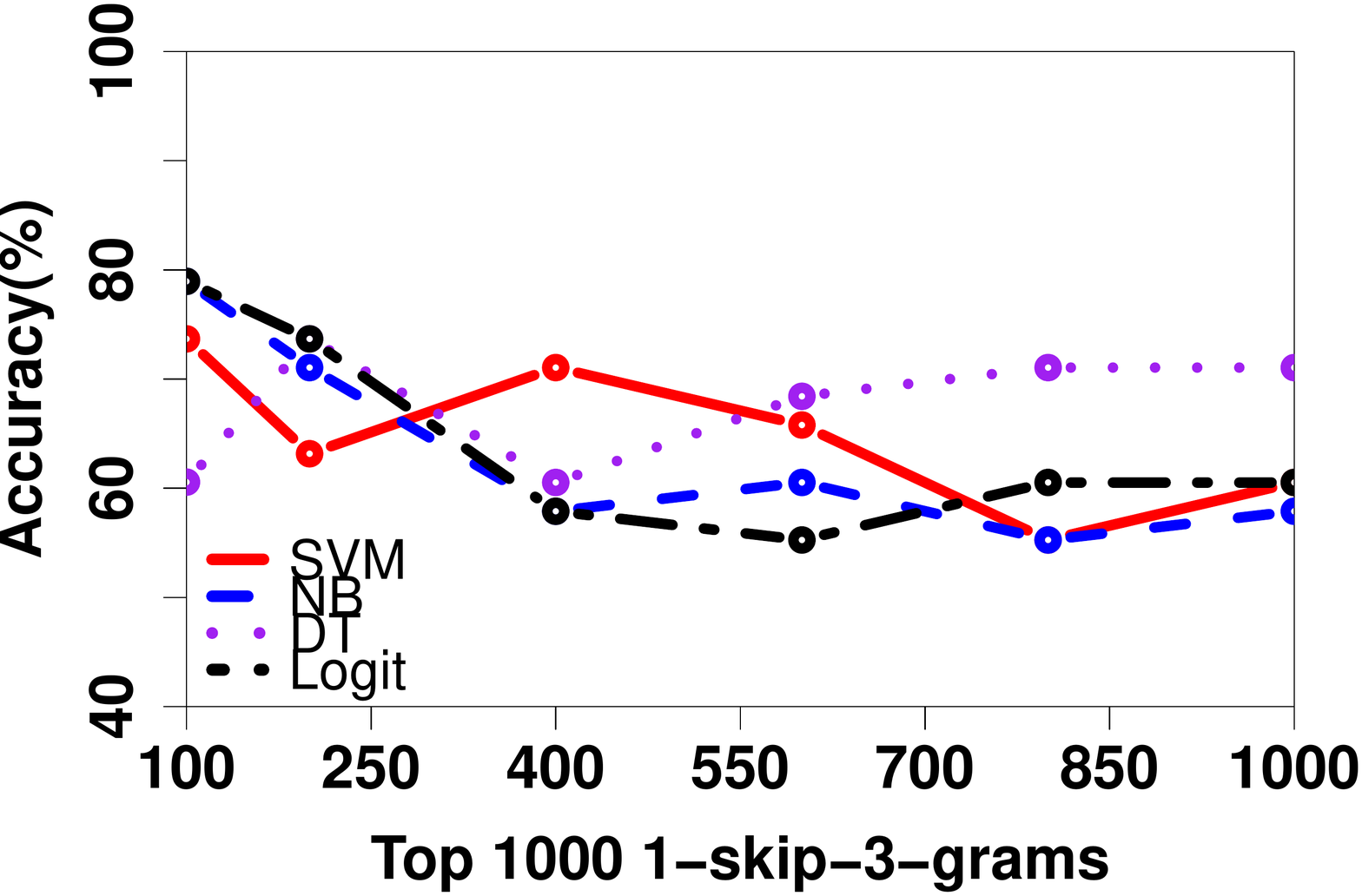}
  \caption{1-skip-3-grams.}\label{fig:1skip3}
\endminipage\hfill
\minipage{0.30\textwidth}%
  \includegraphics[width=40mm,height=40mm]{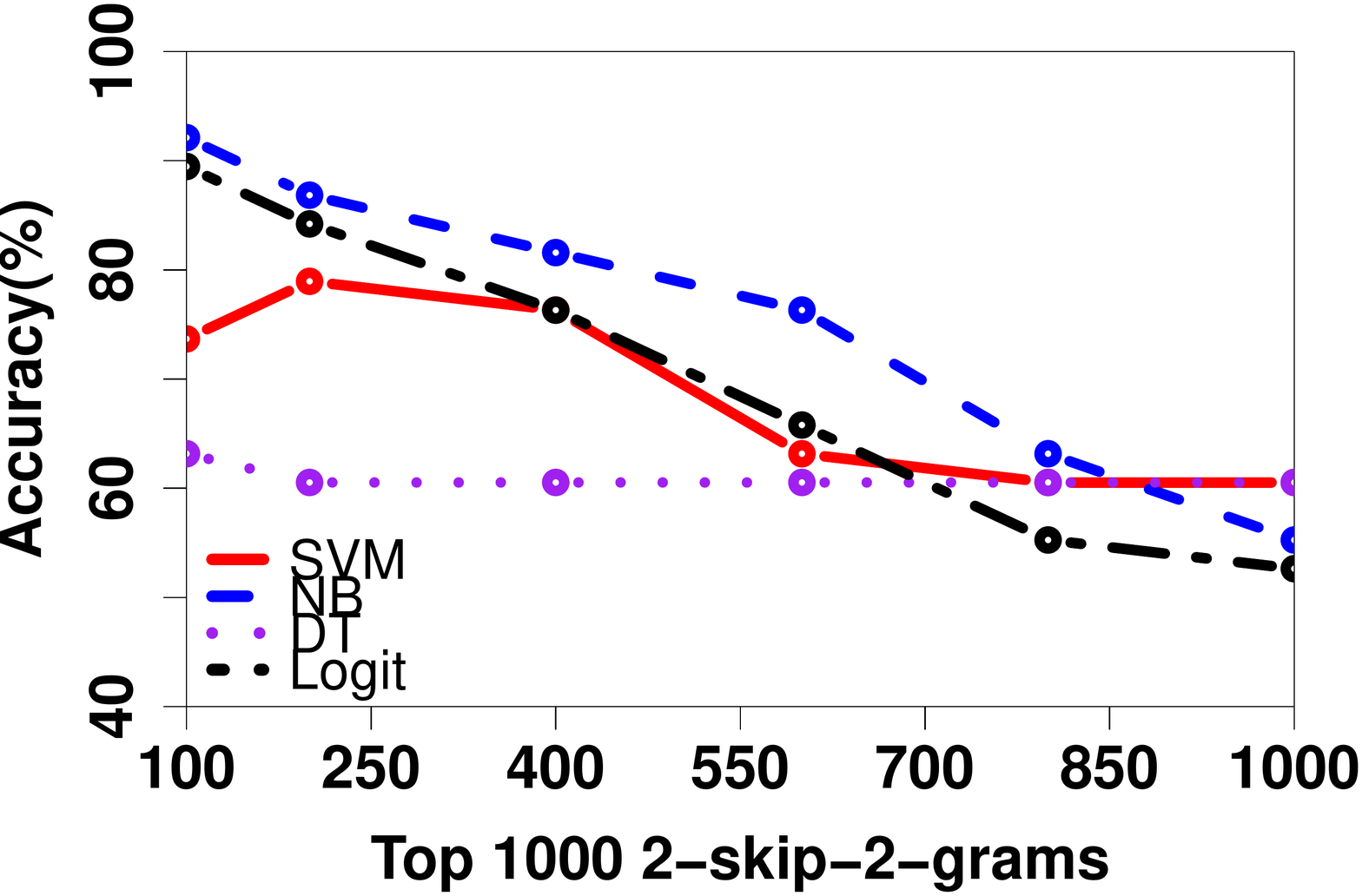}
  \caption{2-skip-2-grams.}\label{fig:2skip2}
\endminipage\hfill
\minipage{0.30\textwidth}%
  \includegraphics[width=40mm,height=40mm]{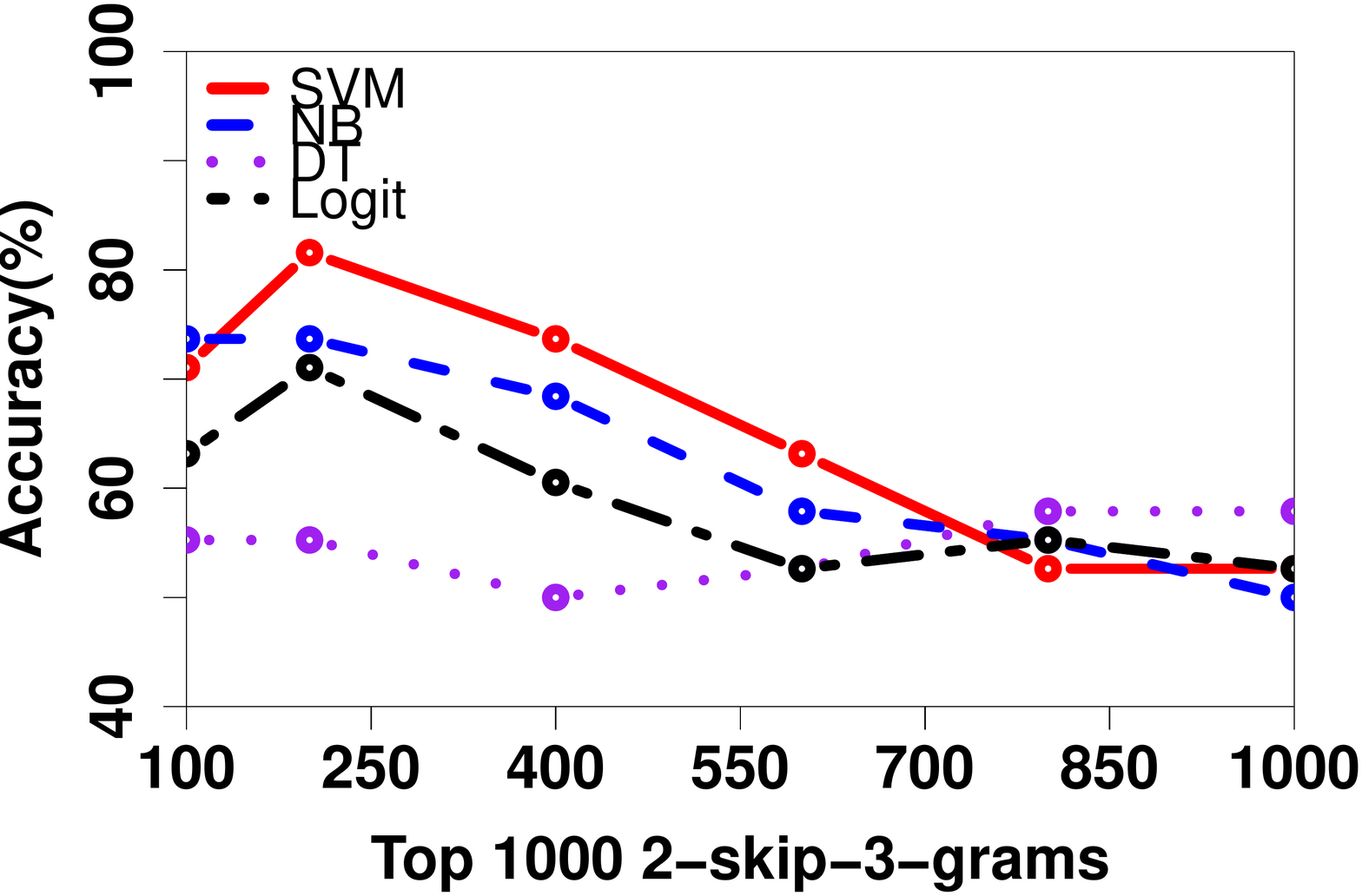}
  \caption{2-skip-3-grams.}\label{fig:2skip3}
\endminipage
\minipage{0.30\textwidth}%
  \includegraphics[width=40mm,height=40mm]{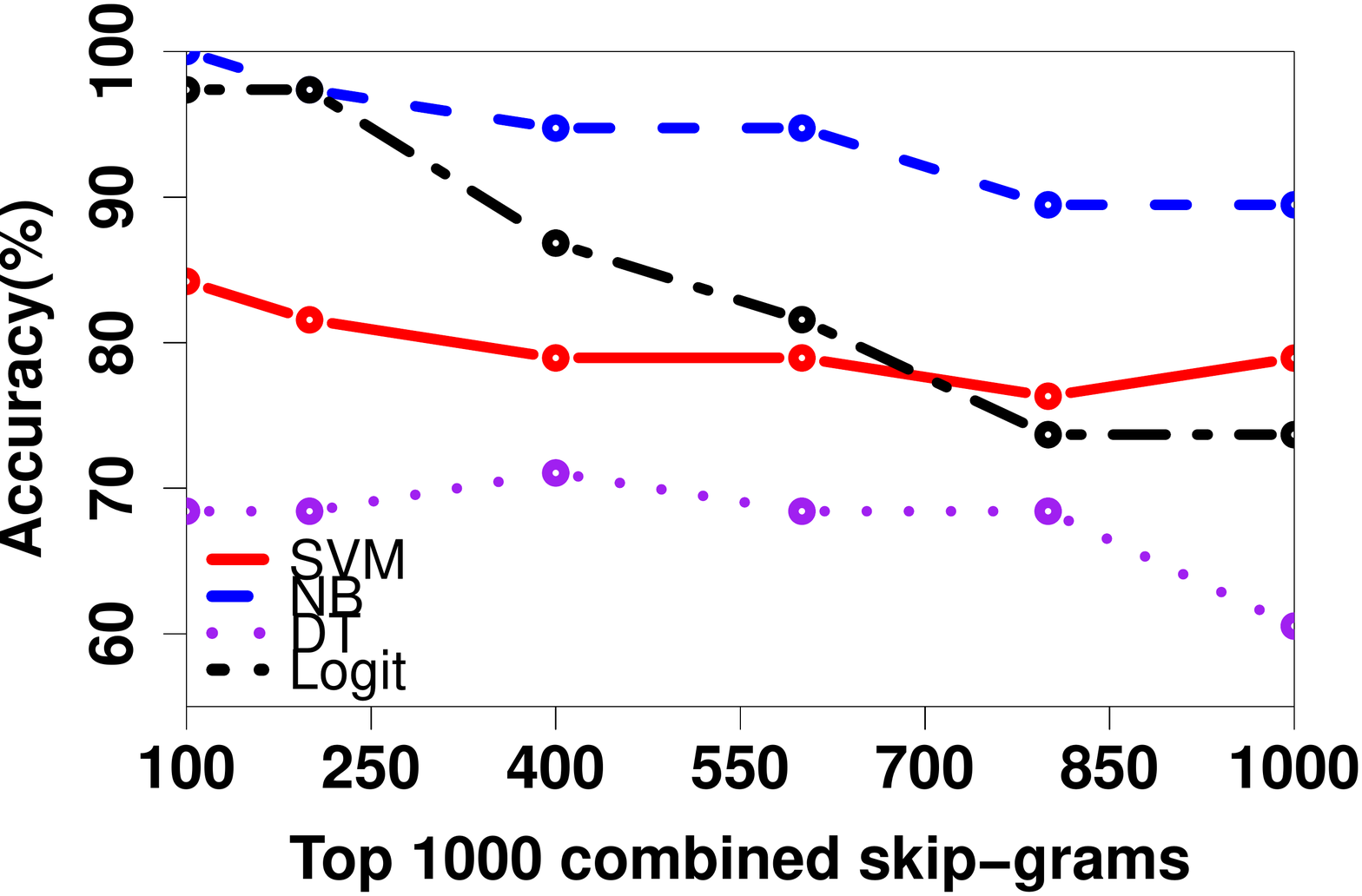}
  \caption{All skip-grams.}\label{fig:allskip}
\endminipage\hfill
\end{figure*}

\begin{table*}[!htb]\footnotesize
\centering
        \begin{tabular}{llllll}
          \hline
          \textbf{Model}  & \textbf{Hyperparameters} & \textbf{Pr.} & \textbf{Rc.} &\textbf{F1} & \textbf{AUC}\\ \hline
SVM & 	-C 0.9375 -N 1 -M -K ``RBFKernel -G 1.0124E-4 & \textbf{0.98} &	\textbf{0.97} &	\textbf{0.97} &	\textbf{0.99} \\
NB	& 	-K & 	\textbf{0.98} &	\textbf{0.97} &	\textbf{0.97} &	\textbf{0.99}\\
DT	&  default (none found by Auto-Weka) &  0.72 &	0.68 &	0.67 &	0.73\\
Logistic &	 -R 8.114737295158544E-12 &	 \textbf{0.98} &	\textbf{0.97} &	\textbf{0.97} &	\textbf{0.99} \\
Baseline-SVM &	-C 0.9681 -N 1 -K NPKernel -E 4.097 -L &	 0.79 &	0.63 &	0.57 &	0.63 \\\hline
        \end{tabular}
        \caption{Top 200 compound skip-grams performance with Auto-Weka hyperparameters.}
        \label{tab:auc}
\end{table*}

\section*{Conclusion and Future Work}
Our results show that linguistic skip-grams could help the diagnosis of MCI. However, these skip-gram features are likely to be limited to the description of the Cookie-Theft picture alone. We plan to conduct further clinical evaluations with leave-pair-out cross-validation and evaluate our models against the MMSE and MoCA on actual MCI patients in the future. 

\subsubsection*{Acknowledgments}
This work was partially funded by the Tropical Medicine and Biology Multidisciplinary Research Platform of Monash University Malaysia and the Malaysian Ministry of Education Fundamental Research Grant Scheme(FRGS) - FRGS/2/2014/ICT07/ MUSM/03/1. This research was conducted with the approval of the Monash University Human Research Ethics Committee (MUHREC) under approval number CF14/240-2014000094.
\small{
\bibliographystyle{ieeetr} 
\bibliography{references} 

\end{document}